\documentclass[runningheads]{llncs}

\usepackage[table,xcdraw,usenames,dvipsnames,svgnames,table]{xcolor}

\usepackage{graphicx}

\usepackage[ruled]{algorithm2e}
\DontPrintSemicolon 

\usepackage{amssymb,amsmath}
\usepackage{amsfonts,braket} 
\usepackage{xspace}

\usepackage[utf8]{inputenc}
\usepackage[english]{babel}
\usepackage[OT2,OT1]{fontenc}

\usepackage{MnSymbol}

\usepackage{tikz}
\usetikzlibrary{arrows,decorations.pathmorphing,fit,positioning,shapes.symbols,chains,shapes.geometric}

\usepackage{url}

\spnewtheorem{defn}{Definition}{\bfseries}{\itshape}

\usepackage{caption}
\usepackage{subcaption}
\newcommand{\rulesep}{\unskip\ \vrule\ }

\usepackage{xargs}

\newcommandx*{\com}[3][2=, 3=blue]{{\color{#3}\textit{[\ldots #1 \ifthenelse{\equal{#2}{}}{}{(#2)}\ldots]}}}

\usepackage{listings}
\usepackage{xcolor}
\usepackage{filecontents}

\makeatletter

\newcommand\PrologPredicateStyle{}
\newcommand\PrologVarStyle{}
\newcommand\PrologAnonymVarStyle{}
\newcommand\PrologAtomStyle{}
\newcommand\PrologOtherStyle{}
\newcommand\PrologCommentStyle{}

\newif\ifpredicate@prolog@
\newif\ifwithinparens@prolog@

\lst@SaveOutputDef{`_}\underscore@prolog

\newcount\currentchar@prolog

\newcommand\@testChar@prolog%
{%
  \ifnum\lst@mode=\lst@Pmode%
    \detectTypeAndHighlight@prolog%
  \else
    \ifwithinparens@prolog@%
      \detectTypeAndHighlight@prolog%
    \fi
  \fi
  \global\predicate@prolog@false%
}

\newcommand\detectTypeAndHighlight@prolog
{%
  \def\lst@thestyle{\PrologAtomStyle}%
  \ifpredicate@prolog@%
    \def\lst@thestyle{\PrologPredicateStyle}%
  \else
    \expandafter\splitfirstchar@prolog\expandafter{\the\lst@token}%
    \expandafter\ifx\@testChar@prolog\underscore@prolog%
      \ifnum\lst@length=1%
        \let\lst@thestyle\PrologAnonymVarStyle%
      \else
        \let\lst@thestyle\PrologVarStyle%
      \fi
    \else
      \currentchar@prolog=65
      \loop
        \expandafter\ifnum\expandafter`\@testChar@prolog=\currentchar@prolog%
          \let\lst@thestyle\PrologVarStyle%
          \let\iterate\relax
        \fi
        \advance \currentchar@prolog by 1
        \unless\ifnum\currentchar@prolog>90
      \repeat
    \fi
  \fi
}
\newcommand\splitfirstchar@prolog{}
\def\splitfirstchar@prolog#1{\@splitfirstchar@prolog#1\relax}
\newcommand\@splitfirstchar@prolog{}
\def\@splitfirstchar@prolog#1#2\relax{\def\@testChar@prolog{#1}}

\def\beginlstdelim#1#2%
{%
  \def\endlstdelim{\PrologOtherStyle #2\egroup}%
  {\PrologOtherStyle #1}%
  \global\predicate@prolog@false%
  \withinparens@prolog@true%
  \bgroup\aftergroup\endlstdelim%
}

\newcommand\lang@prolog{Prolog-pretty}
\expandafter\lst@NormedDef\expandafter\normlang@prolog%
  \expandafter{\lang@prolog}

\expandafter\expandafter\expandafter\lstdefinelanguage\expandafter%
{\lang@prolog}
{%
  language            = Prolog,
  keywords            = {},      
  showstringspaces    = false,
  alsoletter          = (,
  moredelim           = **[is][\beginlstdelim{(}{)}]{(}{)},
  MoreSelectCharTable =
    \lst@DefSaveDef{`(}\opparen@prolog{\global\predicate@prolog@true\opparen@prolog},
}

\newcommand\@ddedToOutput@prolog\relax
\lst@AddToHook{Output}{\@ddedToOutput@prolog}

\lst@AddToHook{PreInit}
{%
  \ifx\lst@language\normlang@prolog%
    \let\@ddedToOutput@prolog\@testChar@prolog%
  \fi
}

\lst@AddToHook{DeInit}{\renewcommand\@ddedToOutput@prolog{}}

\makeatother
\definecolor{PrologPredicate}{RGB}{000,031,255}
\definecolor{PrologVar}      {RGB}{024,021,125}
\definecolor{PrologAnonymVar}{RGB}{000,127,000}
\definecolor{PrologAtom}     {RGB}{186,032,032}
\definecolor{PrologComment}  {RGB}{063,128,127}
\definecolor{PrologOther}    {RGB}{000,000,000}

\renewcommand\PrologPredicateStyle{\color{NavyBlue}}
\renewcommand\PrologVarStyle{\color{Black}}
\renewcommand\PrologAnonymVarStyle{\color{PrologAnonymVar}}
\renewcommand\PrologAtomStyle{\color{DarkGreen}}
\renewcommand\PrologCommentStyle{\itshape\color{DarkGray}}
\renewcommand\PrologOtherStyle{\color{PrologOther}}

\lstdefinestyle{Prolog-pygsty}
{
   language     = Prolog-pretty,
  upquote      = true,
  stringstyle  = \PrologAtomStyle,
  commentstyle = \PrologCommentStyle,
  literate     =
    {:-}{{\PrologOtherStyle :- }}2
    {,}{{\PrologOtherStyle ,}}1
    {.}{{\PrologOtherStyle .}}1
  }

\lstdefinestyle{Prolog-pygsty-sc}
{
  language     = Prolog-pretty,
  upquote      = true,
  stringstyle  = \PrologAtomStyle,
  commentstyle = \PrologCommentStyle,
  basicstyle = \ttfamily,
  literate     =
    {:-}{{\PrologOtherStyle :- }}2
    {,}{{\PrologOtherStyle ,}}1
    {.}{{\PrologOtherStyle .}}1
}

\begin{document}






\title{Explanation as a process: user-centric construction of multi-level and multi-modal explanations
  \thanks{The work presented in this paper is part of the BMBF ML-3 project Transparent
    Medical Expert Companion (TraMeExCo), FKZ~01IS18056~B, 2018-2021.}}

\author{Bettina Finzel  \and David E. Tafler \and Stephan Scheele \and Ute Schmid}

\institute{Cognitive Systems, University of Bamberg 
  \url{https://www.uni-bamberg.de/en/cogsys}
  \email{\{bettina.finzel,stephan.scheele,ute.schmid\}@uni-bamberg.de,  david-elias.tafler@stud.uni-bamberg.de}
}

\titlerunning{Explanation as a process}
\authorrunning{B. Finzel \and D. Tafler \and S. Scheele \and U. Schmid}

\maketitle

\begin{abstract}
  In the last years, XAI research has mainly been concerned with developing new technical
  approaches to explain deep learning models.
  Just
  recent research has started to
  acknowledge the need to tailor explanations to different contexts and requirements of
  stakeholders. Explanations must not only suit developers of models, but also domain experts
  as well as end users. Thus, in order to satisfy different
  stakeholders, explanation methods need to be combined. While multi-modal explanations
  have been used to make model predictions more transparent, less research has focused on
  treating explanation as a process, where users can ask for information according to the
  level of understanding gained at a certain point in time. Consequently, an opportunity
  to explore explanations on different levels of abstraction should be provided besides
  multi-modal explanations. We present a process-based approach that combines multi-level
  and multi-modal explanations. The user can ask for textual explanations or
  visualizations through conversational interaction in a drill-down manner. We use
  Inductive Logic Programming, an interpretable machine learning approach, to learn a
  comprehensible model. Further, we present an algorithm that creates an explanatory tree
  for each example for which a classifier decision is to be explained. The explanatory
  tree can be navigated by the user to get answers of different levels of detail. We
  provide a proof-of-concept implementation for concepts induced from a semantic net about
  living beings.

\end{abstract}

\section{Introduction}
\label{sec:intro}

In order to develop artificial intelligence that serves the human user to perform better
at tasks, it is crucial to make an intelligent system comprehensible to the human user
\cite{miller2019explanation,muggleton_ultra-strong_2018}. This requires giving the user an
understanding of how an underlying algorithm works (mechanistic understanding) on the one
hand and whether the intelligent system fulfills its purpose (functional understanding) on
the other hand \cite{paez_pragmatic_2019}. Explanations can foster both types of
understanding. For example, developers and experts can gain insights into the decision
making process and validate the system. Users who have no in depth technical understanding
in a field could use explainable systems for training, for instance as implemented in
intelligent tutoring systems \cite{putnam_exploring_2019}.

Methods to generate explanations are developed in Explainable Artificial Intelligence
(XAI) research \cite{gunning_darpas_2019}.
A variety of techniques have been proposed,
namely approaches that generate post-hoc explanations for deep learned models
\cite{arrieta2020explainable}, solutions based on interpretable symbolic approaches \cite{rudin2019stop} as well as a
combination of both in hybrid systems \cite{calegari_integration_2020}.
As existing methods are refined and new methods are developed, voices are growing louder about the
need for more user-centric solutions (see for example
\cite{langer_what_2021,paez_pragmatic_2019,rudin2019stop}). Explanations need to fit into
the context of the user, meaning the task and level of expertise. However, there is no
\textit{one-size-fits-all} explanation method, thus approaches have to be combined.

Current research proposes multi-modal explanations to serve the user's varying need for
information \cite{holzinger_towards_2021,el-assady_towards_2019,DBLP:journals/corr/abs-2005-00497}, in particular a
combination of different explanation strategies (inductive, deductive, contrastive) with explanation modalities 
(text, image, audio) to represent information accordingly and involve cognitive processes
adequately for establishing effective XAI methods.
Recent studies show that presenting explanations with different modalities can have a positive influence on the
comprehensibility of a decision. Existing approaches combine visual, textual or
auditory explanations in multi-modal settings \cite{weitz_let_2020,Hendricks_2018_ECCV}.

Less focus is given to explanation as a process. Substantial work exists in the area of
argumentation, machine learning and explanation
\cite{miller2019explanation,walton_dialogue_2016,mozina_argument_2007,hilton_conversational_1991},
where conversations between systems and users follow certain patterns. However, we found
that in the current literature not enough attention is given to the fact that functional
or mechanistic understanding is developed over a period of time and that users may need
different depths and types of information depending on where in the process they are and
which level of expertise they have \cite{el-assady_towards_2019}.
Template-based explanation approaches that allow humans to drill down into an explanation 
and to explore its sub-problems in terms of a hierarchical structure have previously been applied to
assist ontology engineers in understanding inferences and to correct modelling flaws in formal ontologies
\cite{DBLP:journals/ki/LiebigS08} as well to justify results from semantic search \cite{DBLP:journals/ijkedm/Roth-BerghoferF11}.
Studies have shown that users prefer abstract and simple explanations over complex ones
\cite{lombrozo_simplicity_2007}, but may ask for more detailed and complex information
\cite{zemla_evaluating_2017} as well, which should ideally be presented through the best possible and
comprehensible explanation strategy.
Therefore, involving them in a dialogue, where
users can get more detailed explanations if needed, and providing multi-modal explanations
at the same time to reveal different aspects of a prediction and its context, seems to be
a promising step towards more comprehensible decision-making in human-AI-partnerships.

Our contribution, presented in this paper, is an approach that allows users to
understand the classification of a system in a \textit{multi-modal} way and to explore a
system's decision step by step through \textit{multi-level} explanations. Multi-modality
is implemented in a way such that explanations can be requested by the user as textual statements and pictures that illustrate concepts and examples from the domain of interest. Multiple levels of explanation are implemented in such a way that the user can pose three types of requests to the system, which then returns explanations with three different levels of
detail: a \textit{global} explanation to explain a target class, a \textit{local}
explanation to explain the classification of an example with respect to the learned
classification model and \textit{drill-down} explanations that reveal more in-depth
reasons for the classification of an example. Thus, users can ask for different
explanation modalities and details at any point in time in the explanation process
in accordance with their understanding of the system's decision that was gained until this
point in time.

In a proof-of-concept implementation of our proposed approach, we represent a domain of
interest as a semantic net with additional rules to reason about the domain. We train a
model based on Inductive Logic Programming (ILP), since it is a method that allows for
generating interpretable models \cite{rudin2019stop} for relational, potentially highly
complex, domains, where it may be beneficial for the understanding of a user to explain
them in a step-by-step manner. In contrast to other interpretable approaches, ILP allows
for integrating knowledge and learning relations. It has already been applied to concrete
application fields, such as medicine \cite{schmid2020mutual} or more abstract examples,
like family trees \cite{gromowski2020process}, but not yet with a combination of
multi-level and multi-modal explanations. In order to generate these explanations, we
produce a so-called \textit{explanatory tree} for each example that is to be explained.
Each tree is derived from a proof for the classification of an example and extended by
references to images that illustrate concepts in the hierarchy. Finally, we allow for
interaction in natural language with the explanatory tree through a dialogue interface.
Thus, we contribute a concept for multi-level and multi-modal explanations, present an
algorithm to construct such explanations and provide a proof-of-concept implementation
to realize the explanation process accordingly.

The sections of our paper are organized as follows. In Section \ref{sec:example}, we first
describe our running example, i.e., the data basis from a relational domain, which
consists of a semantic net, reasoning rules and examples. In Section \ref{sec:induction},
we show how the chosen ILP approach is used to generate an interpretable model by
generalizing from the data basis. We proceed with describing our approach to multi-level
and multi-modal explanations in Section \ref{sec:multi}. How to generate an explanatory
tree for an example to be explained, is introduced and formalized in Subsection
\ref{sec:trees}. The proof-of-concept implementation is presented in Subsection
\ref{sec:dialogue} and discussed in Subsection \ref{sec:implementation}.
Finally, we show which research aspects and extensions of the system
appear to be of interest for future work in Section \ref{sec:conclusion}.

\section{A Relational Knowledge Domain}
\label{sec:example}

For simple domains, where the model consists just of conjunctions of feature values, it might be sufficient to give a single explanation in one modality. For instance, the prediction that \textit{Aldo plays tennis, if the outlook is sunny and the humidity is normal} (see PlayTennis example from \cite{Mitchell97}) can be plausibly explained by presenting a verbal statement of the learned constraints. In contrast, if a model is more complex, for instance involving relational structures, multi-level and multi-modal explanations might be helpful to foster understanding.

In the following we introduce a running example that we will refer to
throughout the paper to illustrate
our approach.
We assume the reader is familiar with
first-order logic and Prolog (see \cite{sterling_art_1986}), but we will restate the key
terminology. 
The main constructs in Prolog are facts, rules and queries.
Basic Prolog programs consist of a finite set of Horn clauses, where each is a finite
conjunction of literals with at most one positive literal, written in the form 
\begin{align*}
  A_0 \leftarrow A_{1}\wedge A_{2} \wedge \ldots \wedge  A_{n},\ \text{where}\ n \geq 0. 
\end{align*}
Each $A_i$ is an atomic formula of the form $p(t_1,\ldots,t_m)$, consisting of a predicate
$p$ and terms $t_i$, that are either a constant symbol, a variable or a composite
expression. 
The atom $A_0$ is called the \emph{head} and the conjunction of elements
$\bigwedge_{i=1}^n A_i$ is called the \emph{body} of a clause. 
Horn clauses with an empty body are denoted as \emph{facts} and
express unconditional statements, otherwise they are called \textit{rules} and express
conditional statements.
Semantically, we can read a clause as
``the conclusion (or \emph{head}) $A_0$ is true if every $A_i$ in the body is true''.
Facts are literals with constant terms.
Rules express a logical implication to describe that a condition holds if a combination of literals holds,
supported by given facts.
\emph{Queries} can be used to retrieve information from a Prolog program. Queries can be either facts, to check for their truth or falsity or they can be composite expressions to retrieve terms that make those expressions true.
Note that Prolog uses \lstinline!:-! to denote 
$\leftarrow$, 
``\lstinline!,!'' for conjunction $\wedge$ and every clause ends in a full stop. 

Semantic nets are a well-known formalism to represent relational and hierarchical
knowledge. They are constructed from a set of nodes and a set of directed, labeled edges,
where nodes represent concepts and edges denote the semantic relationships between concepts
\cite{chein2008graph,hartley1997semantic}. A semantic network serves as a schema to
represent facts in a declarative way and can therefore be implemented, for example in
Prolog using predicates to represent relations between concepts. 

\begin{figure*}[t!]
\centering
\begin{subfigure}[b]{0.57\textwidth}
         \centering
         \includegraphics[width=\textwidth]{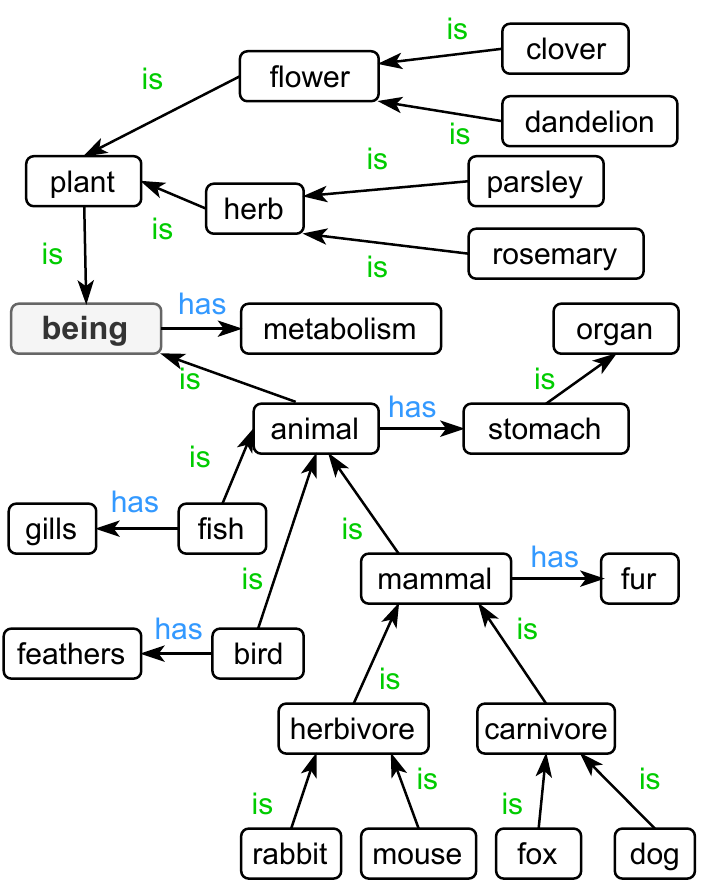}
         \caption{Semantic net}
         \label{sfig:semnet}
       \end{subfigure}
       \enspace
       \rulesep
       \quad
\begin{subfigure}[b]{0.35\textwidth}
\begin{lstlisting}[style=Prolog-pygsty-sc, basicstyle = \fontsize{9}{10}\ttfamily,xleftmargin=.1\textwidth, xrightmargin=.2\textwidth]
is_a(plant,being).
is_a(animal,being).
is_a(flower,plant).
is_a(clover,flower).
is_a(dandelion,flower).
is_a(herb,plant).
is_a(parsley,herb).
is_a(rosemary,herb).
is_a(fish,animal).
is_a(bird,animal).
is_a(mammal,animal).
is_a(herbivore,mammal).
is_a(carnivore,mammal).
is_a(mouse,herbivore).
is_a(rabbit,herbivore).
is_a(fox,carnivore).
is_a(dog,carnivore).
is_a(stomach,organ).

has_p(being,metabolism).
has_p(animal,stomach).
has_p(fish,gills).
has_p(bird,feathers).
has_p(mammal,fur).
\end{lstlisting}
  \caption{Prolog representation}
   \label{sfig:semnetprolog}
\end{subfigure}
\caption{Semantic net of living beings and its representation in Prolog.}
        \label{fig:semnet}
\end{figure*}

Fig.~\ref{sfig:semnet} represents knowledge about living beings and their relations to each
other.
The nodes in the semantic net represent concepts and their subset hierarchy
is expressed through edges via the \lstinline!is! relationship, e.g.,
\lstinline!birds! belong to class \lstinline!animal!. 
The \lstinline!has! relation denotes properties of a concept,
e.g., \lstinline!bird! \emph{has} \lstinline!feathers!.  
Both relations are transitive, e.g., a \lstinline!carnivore! is an
\lstinline!animal!, because it is a \lstinline!mammal! which is an \lstinline!animal!.
Fig.~\ref{sfig:semnetprolog} shows the corresponding
Prolog encoding, where facts consist of a predicate (\lstinline!is_a! or
\lstinline!has_property!) with one or more \emph{constants} as terms, e.g.,
\lstinline!is_a(plant,being)! denotes 
that plant is a 
being.

Reasoning over a Prolog program $P$ is based on the inference rule \textit{modus ponens},
i.e., from $B$ and $A \leftarrow B$ one can deduce $A$,
and the first-order resolution principle and \emph{unification} (the reader may refer to
\cite{bratko_prolog_1986,sterling_art_1986} for more details).
For a query $q$ it verifies whether logical variables can be successfully
substituted with constants from existing facts or some previously implied
conditions from $P$. That means, an existentially quantified query $q$ is a logical
consequence of a program $P$ if there is a clause in $P$ with a ground instance $A
\leftarrow B_{1}, ..., B_{n}, n \geq 0$ such that $B_{1}, ..., B_{n}$ are logical
consequences of $P$, and $A$ is an instance of $q$. Thus, to answer query $A$, the
conjunctive query $\bigwedge_{i=1}^n B_i$ is answered first, since $A$ follows from it. 
For instance, the first reasoning rule from our running example, denoting that
\lstinline!is(A,B) $\;\leftarrow\;$is_a(A,B)! can be used to query the semantic net.
If we pose the query \lstinline!is(animal,being)! accordingly, it will be
\textit{true}, since the predicate \lstinline!is_a(animal,being)! is present in the
facts. With the second reasoning rule \lstinline!is(A,B)! (see below) and the query \lstinline!is(fox,being)! we could, for example, find out that a fox is a living being, due to transitivity.

Additionally, we introduce inheritance and generalization for both relationships, e.g., to  
express that concept rabbit inherits \emph{has} relationships from mammal.
For instance,  a \lstinline!rabbit! \emph{has} \lstinline!fur!, since it \emph{is} a
\lstinline!mammal! that \emph{has} \lstinline!fur!.
Generalisation, on the other hand, allows us to express properties in a
more abstract way, e.g., that an \lstinline!animal! \emph{has} a \lstinline!stomach!
and since the more general concept of \lstinline!stomach! \emph{is} \lstinline!organ!,
it follows that \lstinline!animal! \emph{has} \lstinline!organ!.
Transitivity and inheritance as well as generalisation are expressed by the following reasoning rules:

\begin{lstlisting}[style=Prolog-pygsty-sc, keepspaces=true, showstringspaces=false, showspaces=false,basicstyle = \ttfamily\small , xleftmargin=.05\textwidth, xrightmargin=.2\textwidth]
is(A,B)  :-  is_a(A,B).          is(A,B) :-  is_a(A,C), is(C,B).
has(A,X) :-  has_p(A,X).         has(X,Z) :-  has_p(X,Y), has(Y,Z).
has(A,X) :-  is(A,B), has(B,X).  has(A,X) :-  has_p(A,Y), is(Y,X).
\end{lstlisting}
We will show later how successful unifications resulting from this step-wise reasoning
procedure can be stored in a data structure, which we call \textit{explanatory trees}, to
generate explanations with different levels of detail.

\begin{figure*}[t!]
\centering
\begin{subfigure}[b]{0.35\textwidth}
\begin{lstlisting}[style=Prolog-pygsty-sc, basicstyle = \fontsize{8.5}{9}\ttfamily,  xleftmargin=.05\textwidth, xrightmargin=.1\textwidth]
is_a(bobby,rabbit).
is_a(fluffy,rabbit).
is_a(bella,fox).
is_a(samson,dog).
is_a(argo,dog).
is_a(tipsie,mouse).
is_a(dandelion,flower).
is_a(clover,flower).
is_a(parsley,herb).
is_a(rosemary,herb).
\end{lstlisting}
 \end{subfigure}
        \enspace
       \rulesep
       \quad
\begin{subfigure}[b]{0.45\textwidth}
         \centering
         \includegraphics[width=\textwidth]{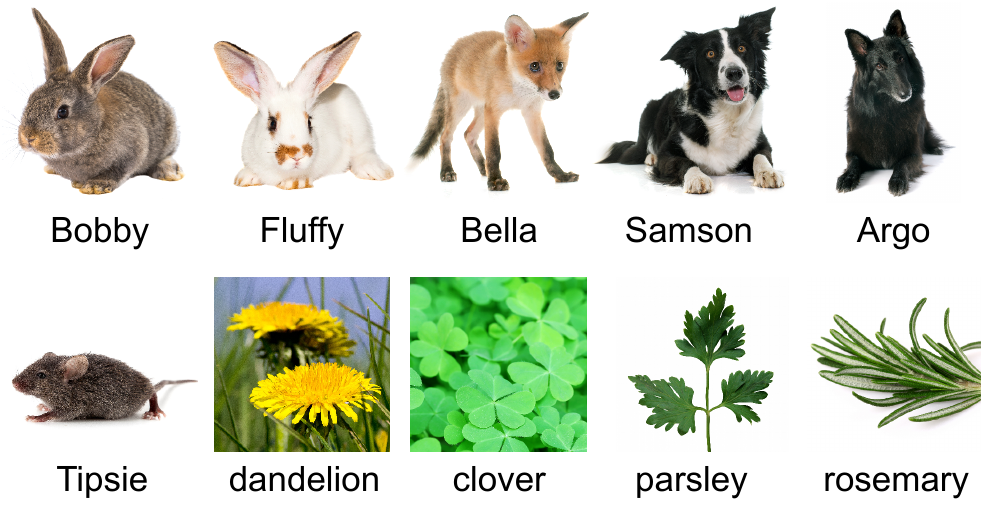}
         \label{sfig:eximg}
       \end{subfigure}
\caption{Background knowledge $T$ with corresponding example images.}
        \label{fig:background}
\end{figure*}

The definition of the semantic net and the reasoning rules represent a so-called background theory $T$ \cite{de_raedt_many_1993}, about living beings
in this case, but it does not include examples yet. In order to represent knowledge about
examples, entries for each example can be added to $T$ by including respective predicates such as shown in Figure~\ref{fig:background}, e.g., to express that
\lstinline!bobby! is an instance of the concept \lstinline!rabbit!.

Having the background theory complemented by knowledge about concrete examples, which
describes relational and hierarchical knowledge, we proceed to take this as input to learn
a classification model for a given target class. From a user's perspective, the target class can be understood as a concept, which we want to explain to the user.
For our running example, we define a relational classification task, where the model has to learn if and under which conditions a living being would
\textit{track down} another living being.

\begin{figure*}[t!]
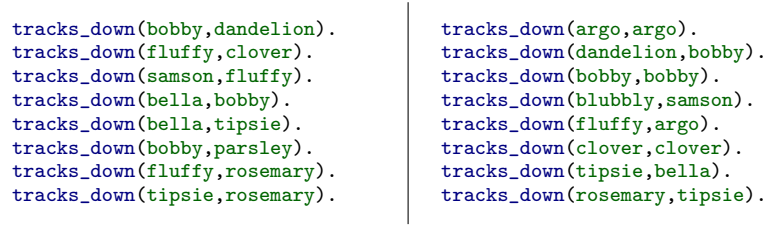

\centering
\begin{subfigure}[b]{0.4\textwidth}
\begin{lstlisting}[style=Prolog-pygsty-sc, basicstyle = \fontsize{8.5}{9}\ttfamily ,  xleftmargin=-.1\textwidth, xrightmargin=.2\textwidth]]
tracks_down(bobby,dandelion).
tracks_down(fluffy,clover).
tracks_down(samson,fluffy).
tracks_down(bella,bobby).
tracks_down(bella,tipsie).
tracks_down(bobby,parsley).
tracks_down(fluffy,rosemary).
tracks_down(tipsie,rosemary).
\end{lstlisting}
 \end{subfigure}
        \qquad\quad
       \rulesep
       \quad
\begin{subfigure}[b]{0.4\textwidth}
\begin{lstlisting}[style=Prolog-pygsty-sc,basicstyle = \fontsize{8.5}{9}\ttfamily,  xleftmargin=.0\textwidth, xrightmargin=.3\textwidth]
tracks_down(argo,argo).
tracks_down(dandelion,bobby).
tracks_down(bobby,bobby).
tracks_down(blubbly,samson).
tracks_down(fluffy,argo).
tracks_down(clover,clover).
tracks_down(tipsie,bella).
tracks_down(rosemary,tipsie).
\end{lstlisting}
 \end{subfigure}
\caption{Positive ($E^+$) and negative ($E^-$) training examples.}
\label{fig:examples}       
\end{figure*}

Consider the set of training examples in Figure \ref{fig:examples}. First, we define a set
$E^+$ of \textit{positive} examples that belong to the target class. For example, the
rabbit \lstinline!bobby! would track down \lstinline!dandelion!.  Secondly, we define a
set $E^-$ of \textit{negative} examples that do not belong to the target class, i.e., we
define cases that exclude reflexivity, inverse transitivity within species and inverses
among all species. For instance, consider the dog \lstinline!argo! that would not track down itself and
obviously also the flower \lstinline!dandelion! would not track down the rabbit \lstinline!bobby!.

\section{Learning an Interpretable Model with ILP}
\label{sec:induction}

 Inductive Logic Programming (ILP) will be applied to induce a classification model in
 terms of inferred rules from the given examples and background theory as introduced
 before.
 For our running example, the model shall learn whether one living being
 would track down another living being.
 ILP is capable of learning a model over hierarchical and non-hierarchical relations that separates positive from
negative examples. The basic learning task of ILP (see e.g., \cite{bratko_prolog_1986} and
\cite{muggleton_inductive_1994}) is described in Algorithm~\ref{fig:algo1} and defined as
follows:

An ILP problem is a tuple $(T,E^+,E^-)$ of ground atoms, where the  background
theory $T$ is a finite set of Horn clauses, and $E^+$ and $E^-$ are disjoint finite sets of positive and negative
example instances.
The goal is to construct a first-order clausal theory $M$ we call \emph{model}, which
is a set of definite clauses that consistently explains the examples w.r.t. the background
theory.
Specifically, it must hold that the model
$M$ is \emph{complete} and \emph{consistent}, i.e.,
\begin{align*}
  \forall p \in E^+\!:\, T \cup M \models p, \text{and}\ \forall  n\in E^-\!:\, T \cup M \not\models n,
\end{align*}
where 
symbol $\models$ denotes the semantic
entailment relation and we assume
the examples to be noise-free, i.e., this excludes false positives and false negatives.

\begin{algorithm}[ht!]
      \caption{Basic ILP algorithm} \label{fig:algo1}
    \textbf{Input:}\;
    \footnotesize $(T,E^+,E^-)$: an ILP problem \;
    \textbf{Output:}\;
    \footnotesize $M$: a classification model \;
    \textbf{Begin:}\;
    \Indp $M \leftarrow \emptyset$, initialise the model\;
    	  $C \leftarrow \emptyset$, temporary clause\; 
    	  $Pos \leftarrow E^{+}$, the set of positive training examples\;
    	  \Indp \textbf{While} $Pos \neq \emptyset$ \textbf{do}\;
				 \Indp $C \leftarrow GenerateNewClause(Pos, E^{-}, T)$\;
				 \Indp	\textbf{such that} $\exists p \in E^+\!:\ T \cup C \models p$\;
				 		\textbf{and} $\forall n \in E^-\!:\ T \cup C \not\models n$\;
				 		\textbf{and} $C$ is optimal w.r.t. quality criterion $A(C)$\;
		  \Indm $Pos \leftarrow Pos \setminus \{p\}$\;
		  		$M \leftarrow M \cup \{C\}$\;
	\Indm \textbf{End}\;
\Indm \textbf{Return} $M$\;
\Indm \textbf{End}
\end{algorithm}

In particular, it can be seen from Algorithm~\ref{fig:algo1} that it learns in a top-down
approach a set of clauses which cover the positive examples while not covering the negative ones.
A model is induced by iteratively generating new clauses $C$ using the function
$GenerateNewClause$~\cite{gromowski2020process} with $T$, $E^+$ and $E{^-}$,
such that for each $C$ there exists a positive
example $p$ that is entailed by $C \cup T$, i.e., $C \cup T \models p$,
while no negative example $n$ is entailed by $C \cup T$.
Furthermore, it is of interest to find clauses such
that each $C$ is optimal with respect to some \textit{quality criterion} $A(C)$, such as the
total number of entailed positive examples.
Such \emph{best} clauses are added to model $M$.

In our approach we use the \textit{Aleph} framework with default settings for the quality criterion \cite{srinivasan_manual} to
generalize clauses (rules) from $(T,E^+,E^-)$.
A well-structured overview on ILP and Aleph in particular is given in Gromowski et al.~\cite{gromowski2020process}.
Given the examples as well as the background theory for our running example, Aleph learns
a model consisting of two rules (see Figure \ref{fig:model}). According to these rules, a
living being $A$ tracks down another living being $B$ either if $A$ is a carnivore and
$B$ is a herbivore, or $A$ is a herbivore and $B$ is a plant.

\begin{figure*}[t!]
\begin{lstlisting}[style=Prolog-pygsty-sc,basicstyle = \ttfamily\footnotesize,  xleftmargin=.15\textwidth, xrightmargin=.1\textwidth]
tracks_down(A,B) :- is(A,carnivore), is(B,herbivore).
tracks_down(A,B) :- is(A,herbivore), is(B,plant). 
\end{lstlisting}
\caption{Learned model}
\label{fig:model} 
\end{figure*}

Note that in our running example, we disregarded carnivorous plants as well as
cannibals. Since ILP works based on the \textit{closed world} assumption, these cases do
not hold given $T$.

\section{Multi-level and Multi-modal Explanations}
\label{sec:multi}

Having the model induced from a relational background theory and training examples, we can now proceed to explain the model’s prediction for an individual example with our proposed multi-level and multi-modal approach that is presented and discussed in the following subsections. We define and explain how our proposed approach generates explanations and how the user can enter into a dialogue with the system to receive them.

\subsection{Explanation Generation}
\label{sec:trees}

Explanations produced by our approach cover three different levels, which can be described as follows.

\begin{itemize}
\item The \textbf{first level} reveals a \textit{global} explanation,
thus an explanation for the target class (e.g., What does \textit{tracks\_down}
mean?). This level of detail gives the most shallow explanation for a target class of
positive examples.
	\item The \textbf{second level} gives a \textit{local} explanation for the
classification of an example by instantiating the global model.
	\item The \textbf{third level} allows for a theoretically endless amount of
detail/drill-down. The drill-down follows each clause's body, which can consist of literals that constitute
heads of further clauses. Thus producing explanations continues as
long as the user asks follow-up questions. However, if the dialogue reaches a fact,
the drill-down ends. The user can then ask
for an image in order to receive a visual explanation.
\end{itemize}

This means, the user can ask for the class of an example, for an explanation for
the class decision as well as ask for explanations for the underlying relational concepts and features in a step-by-step manner.
We define the terminology and different kinds of explanations more formally in the following paragraphs.

Let $P=M\cup T$ be a Prolog program, which we can query to explain the
classification of a positive example $p\in E^+$.
We can create a tree $\varepsilon$ for each clause in $C \in M$ with $C \models p$, which we
call an \textit{explanatory tree}.
This tree is created as part of performing a proof for example $p$, given a clause $C\in
M$ and the corresponding background theory $T$ such as introduced in \cite{bratko_prolog_1986}.
We extend the proof computation by storing $\varepsilon$ consisting
of successful query and sub-query unifications. We check, whether an example
$p \in E^{+}$ can be proven to be entailed by $T \cup C$, where $C \in M$, all generated
by Algorithm~\ref{fig:algo1}.

The tree can be traversed to yield explanations of different levels of detail for
our proposed dialogue approach. In order to generate a \textit{global} explanation
for a target class $c$ (see Def. \ref{def:global-expl}), the set of all clauses from model $M$ of that target class are presented. In order to explain a positive example $p$ globally, only the clauses from $M$ that entail $p$ are presented.
The global explanation corresponds to the body of each clause $C$ from $M$, while the target class is encoded in the heads of all clauses in $M$.
In order to generate a \textit{local} explanation (see Def. \ref{def:local-expl}),
a ground clause $C\theta$ has to be found
for $C$ taken from $M$ or $T$, under substitution $\theta$ of the terms in $C$.
A local explanation is derived from a successful proof of $q$ initialized to the head of $C\theta$,
where the body of the clause $C\theta$ is entailed by $M$ and $T$.
The \textit{drill-down} is applied to a local explanation $C\theta$ (see Def. \ref{def:dd-local-expl}),
and is defined as the creation of a subsequent local explanation for some ground literal $B_{i}\theta$ from
the set of all literals from the body of $C\theta$, given that the body of $C\theta$ is not empty.
If the head of $C\theta$ is a fact and consequently the body of $C\theta$ is empty, the drill-down stops.

\begin{defn}[Global explanation]
  A \emph{global} explanation for a target class $c$
  is given by the set $M$ of all clauses in the learned model.
\label{def:global-expl}
\end{defn}

\begin{defn}[Local explanation]
A \emph{local} explanation for a query $q$ is a ground clause $C\theta$ where $C\in M \cup T$
such that $q=head(C\theta)$ and $M \cup T \models body(C\theta)$.
\label{def:local-expl}
\end{defn}

\begin{defn}[Drill down of a local explanation]
A \emph{drill down} for a local explanation $C\theta$ is given by some literal in
$body(C\theta) = B_1, B_2, \ldots, B_n$ where $n\geq 0$ such that
either $head(C\theta)$ is a fact, i.e., $head(C\theta) \vdash true$ ($body(C\theta) =
\emptyset$);
or otherwise $body(C\theta) \neq \emptyset$ and we create a local explanation for some 
$B_i\theta$, where $1 > i \leq n$. 
\label{def:dd-local-expl}
\end{defn}

Accordingly, the explanatory tree $\varepsilon$ is constructed such that,
for each successful unification of a query $q$,
ground $q$ is the parent node of ground clauses resulting from the proof of $q$.
The explanatory tree $\varepsilon$ can be traversed up and down in a dialogue to get
explanations at all defined levels of detail in the form of natural language expressions
or images at leaf nodes.

\subsection{Explanatory Dialogue}
\label{sec:dialogue}

As presented above, our multi-level and multi-modal explanation approach allows the user to enter into
a dialogue with the system and ask for explanations on three different
levels. Accordingly, users can pose various types of questions, depending on the need
for information and detail.

The input, internal program structure as well as the output of an ILP program is
represented in expressive first-order predicate logic, in Prolog. Nevertheless, although
its output is readable for humans, it is not necessarily comprehensible, due to factors
like the complexity of the domain of interest and the unusual syntax.
We therefore realize the explanatory dialogue by generating natural language expressions.

Template-based generation is commonly used to generate natural language
explanations for logic-based models, e.g., see Siebers et
al. \cite{siebers_please_2019} and Musto et al. \cite{musto_explod_2016}
or \cite{DBLP:journals/ki/LiebigS08}.
Our template consists of domain-specific as well as domain-independent transformation rules.
For example the \textit{has\_a} and \textit{is\_a} relations can be easily translated for
any kind of data set, which is structured as a semantic net. For binary predicates we
connect the first argument with the second argument through a transformed version of the
predicate. In case of a n-ary predicate, arguments are added by \textit{and}. Each parent
node from the explanatory tree is connected to its direct children through the word
\textit{because}, since giving a reason is at the core of implication. Consider the first
rule of our learned model, presented in section \ref{sec:induction}. Transforming this
rule to formulate for example a \textit{global} explanation in natural language results in
the sentence:``A tracks down B, because A is a carnivore and B is a herbivore.''. For a
\textit{local} explanation, the sentence is, e.g.:``Bella tracks down Bobby, because Bella is a
carnivore and Bobby is a herbivore.''.
Beyond this template-based expression generation, our
dialogue provides some static content, such as introductory content, advice and an
epilogue, when the user quits the program.

\begin{figure*}[t!]
  \includegraphics[width=\textwidth]{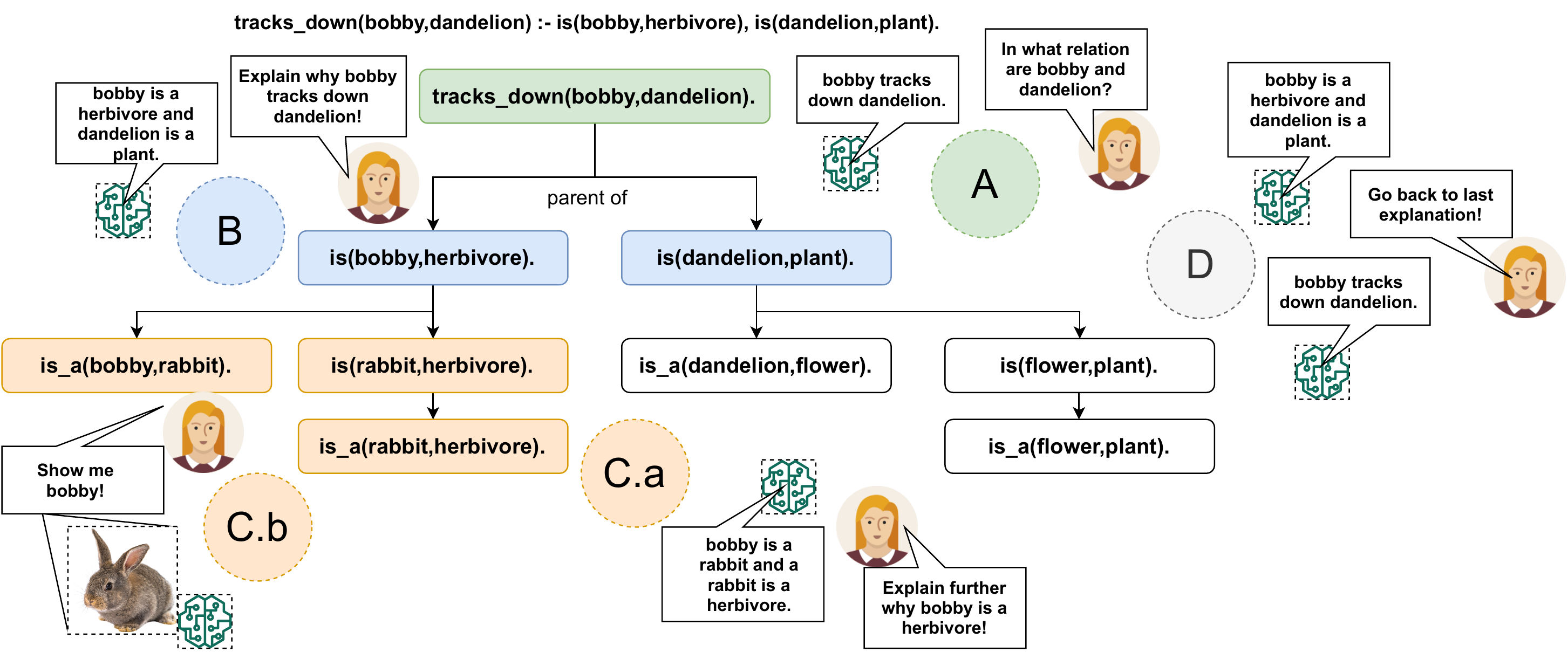}
  \caption{An explanatory tree for \lstinline!tracks_down(bobby,dandelion)!, that can be
    queried by the user to get a \textit{local} explanation why Bobby tracks down dandelion
    (steps A and B). A dialogue is realized by different \textit{drill-down}
    questions, either to get more detailed verbal explanations or visual explanations
    (steps C.a and C.b)). Furthermore, the user can return to the last explanation
    (step D).}
\label{fig:dialogue}   
\end{figure*}

An example of an explanatory dialogue for the
classification of the relationship between \textit{Bobby} and \textit{dandelion} is presented
in Figure \ref{fig:dialogue}. The different levels of explanations correspond to the levels
of the previously generated explanatory tree. The user can explore the whole tree, following
the paths up and down, depending on the type of request (see Figure \ref{fig:dialogue}). Due to space
restrictions we are only presenting the dialogue for local explanations and drill-down requests.
The user could ask for a global explanation by posing the question:``What does tracks\_down mean?''.
The system would return an answer based on the verbalization
of the predicate \lstinline!is! and the constants \lstinline!carnivore! and \lstinline!herbivore! to explain
the first rule of the learned model from Section \ref{sec:induction}.

\subsection{Proof-of-Concept Implementation}
\label{sec:implementation}

The dialogue presented in Fig.~\ref{sec:dialogue} illustrates the proof-of-concept
implementation of our approach. The code is accessible via gitlab\footnote{Gitlab
  repository of the proof-of-concept implementation:
  \url{https://gitlab.rz.uni-bamberg.de/cogsys/public/multi-level-multi-modal-explanation}}.
In general, the approach is not only applicable to the domain presented here. Any Prolog
program including a model learned with ILP can be explained with our approach. The only
precondition is, that the template-based verbalization of the nodes from the explanatory
tree can be performed by the domain-independent transformation rules.
Otherwise, domain-specific transformation rules must be defined first.
Furthermore, we want to point out that the introduced algorithm fulfills correctness,
since it is complete and consistent with respect to the ILP problem solution.  Finally,
first empirical investigations show that humans perform better in decision-making and
trust more into the decisions of an intelligent system, if they are presented with visual
as well as verbal explanations \cite{thaler_schmid}.

\section{Conclusion and Outlook}
\label{sec:conclusion}
Explanations of decisions made by intelligent systems need to be tailored to the needs of
different stakeholders. At the same time there exists no \textit{one-size-fits-all}
explanation method. Consequently, an approach that combines explanation modalities and
that provides explanation as a step-by-step process, is promising to satisfy various
users. We presented an approach that combines textual and visual explanations, such that
the user can explore different kinds of explanations by posing requests to the system
through a dialogue. We introduced an algorithm as well as our proof-of-concept
implementation of it.

In comparison to other explainable state-of-the-art systems that present explanations
rather as static content \cite{arrieta2020explainable}, our approach allows for step-by-step exploration of the reasons
behind a decision of an intelligent system. Since our approach is interpretable, it could
help users in the future to uncover causalities between data and a system's prediction.
This is especially important in decision-critical areas, such as medicine
\cite{holzinger_causability_2019,bruckert2020next,schmid2020mutual}.

Other recent interactive systems enable the user to perform
corrections on labels and to act upon wrong explanations, such as implemented in the CAIPI approach
\cite{teso2019explanatory}, they allow for re-weighting of features for explanatory
debugging, like the EluciDebug system \cite{kulesza2010explanatory} and correcting generated
verbal explanations and the underlying model through user-defined constraints, such as
implemented in the medical-decision support system LearnWithME \cite{schmid2020mutual}.
Our approach could be extended by correction capabilities in the future, in addition to requesting
information from the system to better understand its operation or
purpose. In that way, explanations would be bi-directional.

We presented a proof-of-concept implementation of our approach that could be
technically extended in the future by combining explanations with linked data, e.g.,
to integrate formal knowledge from ontologies combined with media from open data
repositories, which would allow for more flexibility in the presentation of content based
on semantic search\footnote{Retrieval through semantic search can be performed for example
  over semantic \textit{mediawiki}:
  https://www.semantic-mediawiki.org/wiki/Help:Semantic\_search}.
Furthermore, we envisage to explain decisions of deep neural networks using ILP, as presented in \cite{rabold_explaining_2018}.

In future work, we aim to systematically evaluate our approach with empirical user studies, recognizing design dimensions of XAI put forth by Sperrle et al. \cite{sperrle_should_2020}. In this context, our multi-modal and multi-level approach allows for precise control and manipulation of experimental conditions. For instance, we are currently preparing an empirical study to evaluate the effectiveness of the combination of multi-level and multi-modal explanations with respect to user performance and understanding of a model. It is planned to apply the proposed approach to a decision-critical scenario, in particular to asses pain and emotions in human facial expressions for a medical use case.

In order to meet a user-centric explanation design and to close the semantic gap between
data, model and explanation representation, the use of hybrid approaches seems promising
as a future step. Especially in the field of image-based classification, it is an exciting
task to consider sub-symbolic neural networks as sensory components and to combine
them with symbolic approaches as drivers of knowledge integration. In this way, models can
be validated and corrected with the help of expert knowledge. Furthermore, hybrid
approaches allow for modeling the context of an application domain in a more expressive
and complex manner and, in addition, taking into account users' context and individual differences. Our
approach is a first step towards this vision.
%
%
%
\medskip

\noindent
\textbf{Acknowledgements:}
The authors would like to thank the anonymous referees, who provided useful comments on the submission version of the paper.


\bibliographystyle{splncs04}
\bibliography{multi-level}   


\end{document}